\documentclass{article}

\usepackage{nips13submit_e,times}
\usepackage{amsmath,amsfonts,amssymb}
\usepackage{booktabs}
\usepackage{epsfig}
\usepackage{graphicx}
\usepackage{verbatim}
\usepackage{amsmath}
\usepackage{textcomp}
\usepackage{amssymb}
\usepackage{url}
\usepackage{chngpage}

\usepackage{amsmath,amsfonts,amssymb}
\usepackage{multirow}
\usepackage{chngpage}

\usepackage{marginnote}
\usepackage{paralist}
\usepackage[normalem]{ulem}
\usepackage{sidecap}
\usepackage{float}
\usepackage{caption3}
\DeclareCaptionOption{parskip}[]{}
\usepackage[small]{caption}
\usepackage{rotating}
\usepackage{subfigure}

\DeclareMathOperator*{\argmax}{arg\,max}

\usepackage{hyperref}
\usepackage{xcolor}
\definecolor{dark-red}{rgb}{0.4,0.15,0.15}
\definecolor{dark-blue}{rgb}{0.15,0.15,0.4}
\definecolor{medium-blue}{rgb}{0,0,0.5}
\hypersetup{
    colorlinks, linkcolor={black},
    citecolor={dark-blue}, urlcolor={medium-blue}
}

\title{Intriguing properties of neural networks}

\author{
Christian Szegedy\\
Google Inc.\\
\And
Wojciech Zaremba\\
New York University\\ 
\And
Ilya Sutskever\\
Google Inc.\\
\And
Joan Bruna\\
New York University\\
\And
Dumitru Erhan\\
Google Inc.\\
\And
Ian Goodfellow\\
University of Montreal\\
\And
Rob Fergus\\
New York University\\
Facebook Inc.\\
}

\nipsfinalcopy 

\begin{document}

\maketitle

\newtheorem{theorem}{Theorem}[section]
\newtheorem{lemma}[theorem]{Lemma}
\newtheorem{proposition}[theorem]{Proposition}
\newtheorem{corollary}[theorem]{Corollary}

\definecolor{LimeGreen}{rgb}{0.1,0.9,0.1}
\definecolor{Maroon}{rgb}{0.9,0.1,0.1}

\begin{abstract}

Deep neural networks are highly expressive models that have recently
achieved state of the art performance on speech and visual recognition
tasks.  While their expressiveness is the reason they succeed, it also causes
them to learn uninterpretable solutions that could have counter-intuitive properties. 
In this paper we report two such properties.

First, we find that there is no distinction between individual high level units and random linear 
combinations of high level units, according to various methods of unit analysis. It suggests
that it is the space, rather than the individual units, that contains the semantic information 
in the high layers of neural networks.

Second, we find that deep neural networks learn input-output mappings that are fairly discontinuous
to a significant extent.
We can cause the network to misclassify an image by applying a certain
hardly perceptible perturbation, which is found by maximizing the network's prediction error.
In addition, the specific nature of these perturbations is not a random artifact of learning:  the same perturbation     
can cause a different network, that was trained on a different subset of the dataset, to misclassify the same input.

\end{abstract}

\section{Introduction}

Deep neural networks are powerful learning models that achieve
excellent performance on visual and speech recognition problems
\cite{krizhevsky2012imagenet, deepSpeechReviewSPM2012}. 
Neural networks achieve high performance because they can express
arbitrary computation that consists of a modest number of massively parallel
nonlinear steps. But as the resulting computation is automatically discovered by backpropagation via supervised
learning, it can be difficult to interpret and can have counter-intuitive properties. 
In this paper, we discuss two counter-intuitive properties of deep neural networks.

The first property is concerned with the semantic meaning of individual units. Previous
works \cite{girshick2013rich, zeiler2013visualizing, goodfellow2009measuring} analyzed
the semantic meaning of various units by finding the set of inputs
that maximally activate a given unit.  The inspection of individual units makes the implicit
assumption that the units of the last feature layer form a distinguished basis which is
particularly useful for extracting semantic information.
Instead, we show in section \ref{sec:image} that random projections of $\phi(x)$ are semantically
indistinguishable from the coordinates of $\phi(x)$. This puts into question 
the conjecture that neural networks disentangle variation factors across coordinates.
Generally, it seems that it is the entire space of activations,
rather than the individual units, that contains the bulk of the semantic information.
A similar, but even stronger conclusion was reached recently by Mikolov et al.
\cite{mikolov2013efficient} for word representations, where the various directions
in the vector space representing the words are shown to give rise to a
surprisingly rich semantic encoding of relations and analogies.
At the same time, the vector representations are stable up to a rotation of the
space, so the individual units of the vector representations are unlikely to
contain semantic information.

The second property is concerned with the stability of neural networks with respect to small
perturbations to their inputs. Consider a state-of-the-art
deep neural network that generalizes well on an object recognition task. 
We expect such  network to be robust to small perturbations of its input, 
because small perturbation cannot change the object category of an image.
However, we find that applying an \emph{imperceptible}
non-random perturbation to a test image, it is possible to arbitrarily change
the network's prediction  (see figure \ref{fig:alexnegative}).
These perturbations are found by optimizing the input to maximize
the prediction error.  We term the so perturbed examples ``adversarial examples''.

It is natural to expect that the precise configuration of the minimal
necessary perturbations is a random artifact of the normal variability that
arises in different runs of backpropagation learning. Yet, we found that
adversarial examples are relatively robust, and are shared by neural networks
with varied number of layers, activations or trained on different subsets
of the training data.
That is, if we use one neural net to generate a set of adversarial examples, we find
that these examples are still statistically hard for another neural network
even when it was trained with different hyperparameters or, most surprisingly,
when it was trained on a different set of examples.

These results suggest that the deep neural networks that are learned by
backpropagation have nonintuitive characteristics and intrinsic blind spots, whose structure
is connected to the data distribution in a non-obvious way.

\section{Framework}

{\bf Notation} We denote by $x \in \mathbb{R}^m$
an input image, and $\phi(x)$ activation values of some layer. We first examine properties
of the image of $\phi(x)$, and then we search for its blind spots.

We perform a number of experiments on a few different networks and three datasets :
\begin{itemize}
	\item For the MNIST dataset, we used the following architectures~\cite{lecun1998mnist}
	\begin{itemize}
		\item A simple fully connected network with one or more hidden layers and a Softmax classifier. We refer to this network as ``FC''.
		\item A classifier trained on top of an autoencoder. We refer to this network as ``AE''.
	\end{itemize}
	\item The ImageNet dataset~\cite{deng2009imagenet}. 
	\begin{itemize}
		\item Krizhevsky et. al architecture \cite{krizhevsky2012imagenet}. We refer to it as ``AlexNet''.
	\end{itemize}
	\item $\sim 10$M image samples from Youtube (see~\cite{le2011building})
	\begin{itemize}
		\item Unsupervised trained network with $\sim$ 1 billion learnable parameters. We refer to it as ``QuocNet''.
	\end{itemize}
\end{itemize}

For the MNIST experiments, we use regularization with a weight decay of $\lambda$. Moreover, in some experiments we 
split the MNIST training dataset into two disjoint datasets $P_1$, and $P_2$, each with 30000 training cases.

\section{Units of: $\phi(x)$}\label{sec:image}

Traditional computer vision systems rely on feature extraction:
often a single feature is easily interpretable, e.g. a histogram of colors, or quantized local derivatives. This allows one to
inspect the individual coordinates of the feature space, and link them back to meaningful variations in the input domain.
Similar reasoning was used in previous work that attempted to analyze neural networks that were applied to computer vision problems. 
These works interpret an activation of a hidden unit as
a meaningful feature. They look for input images which maximize the activation value of this single feature~\cite{girshick2013rich, zeiler2013visualizing, goodfellow2009measuring,erhan2009visualization}.

The aforementioned technique can be formally stated as visual inspection of images $x'$, which satisfy (or are close to maximum attainable value): 
\begin{align*}
	x' = \argmax_{x \in \mathcal{I}} \langle \phi(x), e_i \rangle
\end{align*}
where $\mathcal{I}$ is a held-out set of images from the data distribution that the network
was not trained on and $e_i$ is the natural basis vector associated with the $i$-th hidden unit.

Our experiments show that any random direction $v \in \mathbb{R}^n$ gives rise to similarly interpretable
semantic properties. More formally, we find that images $x'$ are semantically related to each
other, for many $x'$ such that
\begin{align*}
	x' = \argmax_{x \in \mathcal{I}} \langle \phi(x), v \rangle
\end{align*}

This suggests that the natural basis is not better than a random basis for inspecting the properties of $\phi(x)$. This puts into question the notion that neural networks disentangle variation factors across coordinates.

\begin{figure}
  \centering
  \subfigure[Unit sensitive to lower round stroke.]{
        \includegraphics[width=0.37\textwidth]{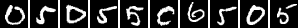}
  }
  \hspace{5mm}
  \subfigure[Unit sensitive to upper round stroke, or lower straight stroke.]{
        \includegraphics[width=0.37\textwidth]{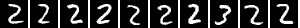}
  }
  \hspace{5mm}
  \subfigure[Unit senstive to left, upper round stroke.]{
        \includegraphics[width=0.37\textwidth]{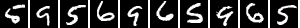}
  }
  \hspace{5mm}
  \subfigure[Unit senstive to diagonal straight stroke.]{
        \includegraphics[width=0.37\textwidth]{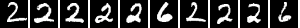}
  }
  \caption{An MNIST experiment. The figure shows images that maximize the activation of various units (maximum stimulation in the
natural basis direction). Images within each row share semantic properties.}
  \label{fig:natural}
\end{figure}

\begin{figure}
  \centering
  \subfigure[Direction sensitive to upper straight stroke, or lower round stroke.]{
        \includegraphics[width=0.37\textwidth]{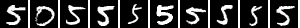}
  }
  \hspace{5mm}
  \subfigure[Direction sensitive to lower left loop.]{
        \includegraphics[width=0.37\textwidth]{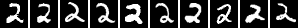}
  }
  \hspace{5mm}
  \subfigure[Direction senstive to round top stroke.]{
        \includegraphics[width=0.37\textwidth]{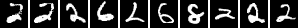}
  }
  \hspace{5mm}
  \subfigure[Direction sensitive to right, upper round stroke.]{
        \includegraphics[width=0.37\textwidth]{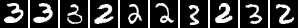}
  }
  \caption{An MNIST experiment. The figure shows images that maximize the activations in a random direction
(maximum stimulation in a random basis). Images within each row share semantic properties.}
  \label{fig:random}
\end{figure}

First, we evaluated the above claim using a convolutional neural network trained on MNIST.
We used the MNIST test set for $\mathcal{I}$.
Figure \ref{fig:natural}
shows images that maximize the activations in the natural basis, and Figure \ref{fig:random} shows images
that maximize the activation in random directions. In both cases the resulting images share many high-level similarities. 

\begin{figure}
  \centering
  \subfigure[Unit sensitive to white flowers.]{
        \includegraphics[width=0.37\textwidth]{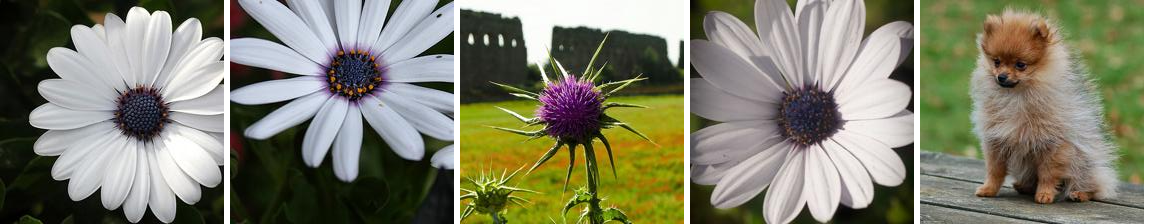}
  }
  \hspace{5mm}
  \subfigure[Unit sensitive to postures.]{
        \includegraphics[width=0.37\textwidth]{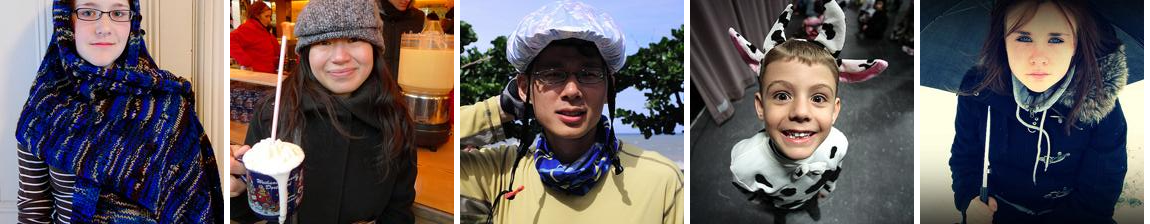}
  }
  \hspace{5mm}
  \subfigure[Unit senstive to round, spiky flowers.]{
        \includegraphics[width=0.37\textwidth]{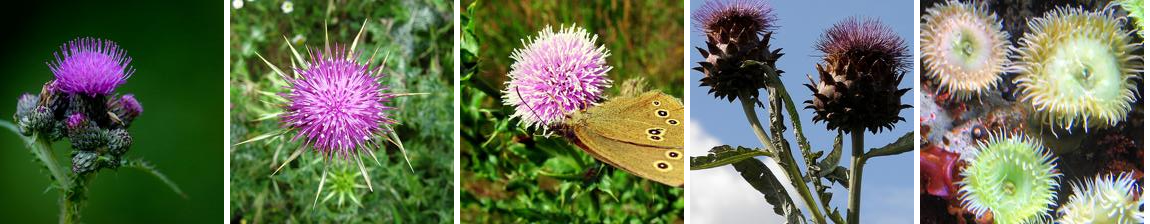}
  }
  \hspace{5mm}
  \subfigure[Unit senstive to round green or yellow objects.]{
        \includegraphics[width=0.37\textwidth]{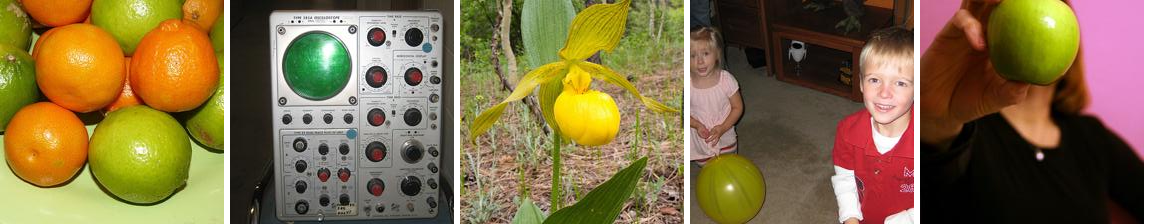}
  }
  \caption{Experiment performed on ImageNet. Images stimulating single unit most (maximum stimulation in natural basis direction). Images within each row share many semantic properties.}
  \label{fig:natural_imagenet}
\end{figure}

\begin{figure}
  \centering
  \subfigure[Direction sensitive to white, spread flowers.]{
        \includegraphics[width=0.37\textwidth]{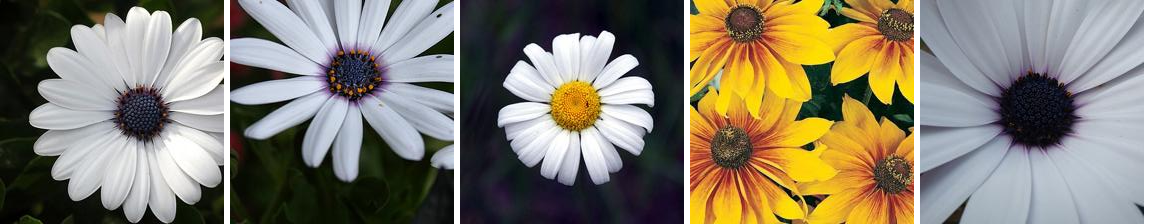}
  }
  \hspace{5mm}
  \subfigure[Direction sensitive to white dogs.]{
        \includegraphics[width=0.37\textwidth]{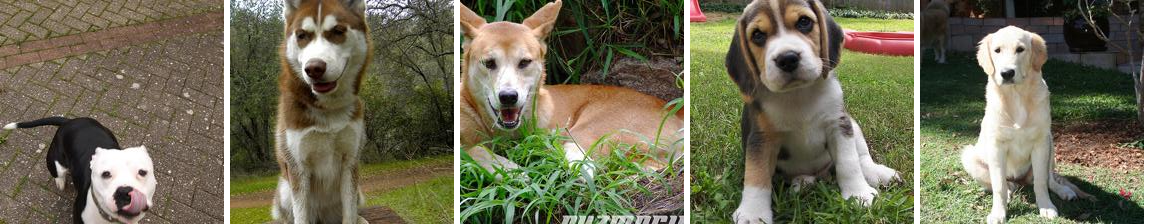}
  }
  \hspace{5mm}
  \subfigure[Direction sensitive to spread shapes.]{
        \includegraphics[width=0.37\textwidth]{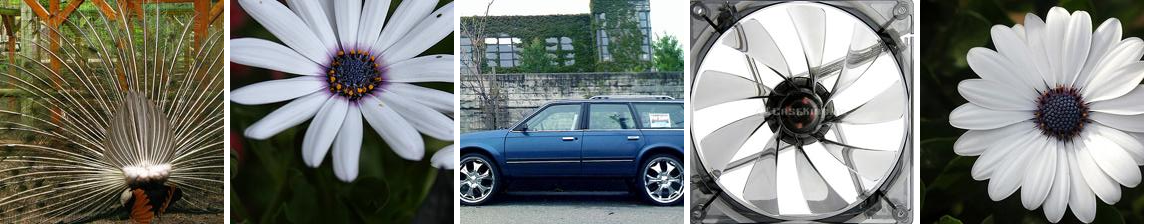}
  }
  \hspace{5mm}
  \subfigure[Direction sensitive to dogs with brown heads.]{
        \includegraphics[width=0.37\textwidth]{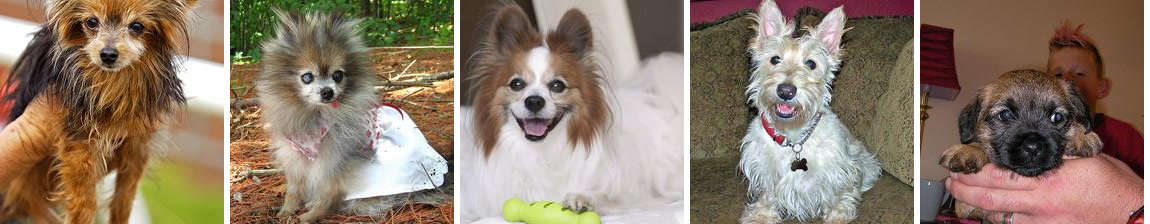}
  }
  \caption{Experiment performed on ImageNet. Images giving rise to maximum activations in a random direction (maximum stimulation in a random basis). Images within each row share many semantic properties.}
  \label{fig:random_imagenet}
\end{figure}

Next, we repeated our experiment on an AlexNet, where we used the validation set as $\mathcal{I}$. Figures \ref{fig:natural_imagenet} and \ref{fig:random_imagenet} compare the natural basis to the random basis on the trained network. The rows appear to be semantically meaningful for both the single unit and the combination of units.

Although such analysis gives insight on the capacity of $\phi$ to generate invariance on a particular subset of the input
distribution, it does not explain the behavior on the rest of its domain. We shall see in the next section that $\phi$ has
counterintuitive properties in the neighbourhood of almost every point form data distribution.


\section{Blind Spots in Neural Networks}\label{sec:discont}

So far, unit-level inspection methods had relatively little utility beyond confirming certain intuitions regarding the complexity of the representations learned by a deep neural network~\cite{girshick2013rich, zeiler2013visualizing, goodfellow2009measuring,erhan2009visualization}. Global, network level inspection methods \emph{can} be useful in the context of explaining classification decisions made by a model~\cite{baehrens2010explain} and can be used to, for instance, identify the parts of the input which led to a correct classification of a given visual input instance
(in other words, one can use a trained model for weakly-supervised localization). Such global analyses are useful in that they can make us understand better the input-to-output mapping represented by the trained network.

Generally speaking, the output layer unit of a neural network is a highly nonlinear function of its input. When it is trained with the cross-entropy loss (using the Softmax activation function), it represents a conditional distribution of the label given the input (and the training set presented so far). It has been argued~\cite{bengio2009learning} that the deep stack of non-linear layers in between the input and the output unit of a neural network are a way for the model to encode a \emph{non-local generalization prior} over the input space.
In other words, it is assumed that is possible for the output unit to assign non-significant
(and, presumably, non-epsilon) probabilities to regions of the input space that contain no training examples in their vicinity. Such regions can represent, for instance, the same objects from different viewpoints, which are relatively far (in pixel space), but which share nonetheless both the label and the statistical structure of the original inputs.

It is implicit in such arguments that $local$ generalization---in the very proximity of the training examples---works as expected. And that in particular, for a small enough radius $\varepsilon > 0$ in the vicinity of a given training input $x$, an $x + r$ satisfying $||r|| < \varepsilon$ will get assigned a high probability of the correct class by the model. This kind of smoothness prior is typically valid for computer vision problems. In general, imperceptibly tiny perturbations of a given image do not normally change the underlying class.


Our main result is that for deep neural networks, the smoothness assumption that underlies many kernel methods does not hold.  Specifically, we show that by using a simple optimization procedure, we are able to find adversarial examples, which are obtained by imperceptibly small perturbations to a correctly classified input image, so that it is no longer classified correctly.

In some sense, what we describe is a way to traverse the manifold represented by the network in an efficient way (by optimization) and finding \emph{adversarial examples} in the input space. The adversarial examples represent low-probability (high-dimensional) ``pockets'' in the manifold, which are hard to efficiently find by simply randomly sampling the input around a given example. Already, a variety of recent state of the art computer vision models employ input deformations during training for increasing the robustness and convergence speed of the models~\cite{krizhevsky2012imagenet,zeiler2013visualizing}. These deformations are, however, statistically inefficient, for a given example: they are highly correlated and are drawn from the same distribution throughout the entire training of the model. We propose a scheme to make this process adaptive in a way that exploits the model and its deficiencies in modeling the local space around the training data.

We make the connection with hard-negative mining explicitly, as it is close in spirit: hard-negative mining, in computer vision, consists of identifying training set examples (or portions thereof) which are given low probabilities by the model, but which should be high probability instead, cf.~\cite{felzenszwalb2008discriminatively}. The training set distribution is then changed to emphasize such hard negatives and a further round of model training is performed. As shall be described, the optimization problem proposed in this work can also be used in a constructive way, similar to the hard-negative mining principle.

\subsection{Formal description}
We denote by $f: \mathbb{R}^m \longrightarrow \{1 \dots k\}$ a classifier
mapping image pixel value vectors to a discrete label set. We also assume
that $f$ has an associated continuous loss function denoted by
$\textrm{loss}_f:\mathbb{R}^m\times \{1 \dots k\}\longrightarrow \mathbb{R}^+$.
For a given $x\in\mathbb{R}^m$ image and target label $l\in \{1 \dots k\}$,
we aim to solve the following box-constrained optimization problem:
\begin{itemize}
\item Minimize $\|r\|_2$ subject to:
  \begin{enumerate}
    \item $f(x+r)=l$
    \item $x+r \in [0, 1]^m$
  \end{enumerate}
\end{itemize}
The minimizer $r$ might not be unique, but we denote one such $x+r$ for an
arbitrarily chosen minimizer by $D(x, l)$. Informally, $x + r$ is the closest
image to $x$ classified as $l$ by $f$.
Obviously, $D(x, f(x))=f(x)$, so this task is non-trivial only if
$f(x)\neq l$.
In general, the exact computation of $D(x, l)$ is a hard problem,
so we approximate it by using a box-constrained
L-BFGS. Concretely, we find an approximation of $D(x, l)$ by performing
line-search to find the minimum $c>0$ for which the minimizer $r$
of the following problem satisfies $f(x+r)=l$.
\begin{itemize}
\item Minimize $c|r| + \textrm{loss}_f(x + r, l)$ subject to $x+r \in [0,1]^m$
\end{itemize}
This penalty function method would yield the exact solution for $D(X, l)$
in the case of convex losses, however neural networks are non-convex in general,
so we end up with an approximation in this case.
\subsection{Experimental results}
\begin{figure}
  \centering
  \subfigure[] {
        \includegraphics[width=0.32\textwidth]{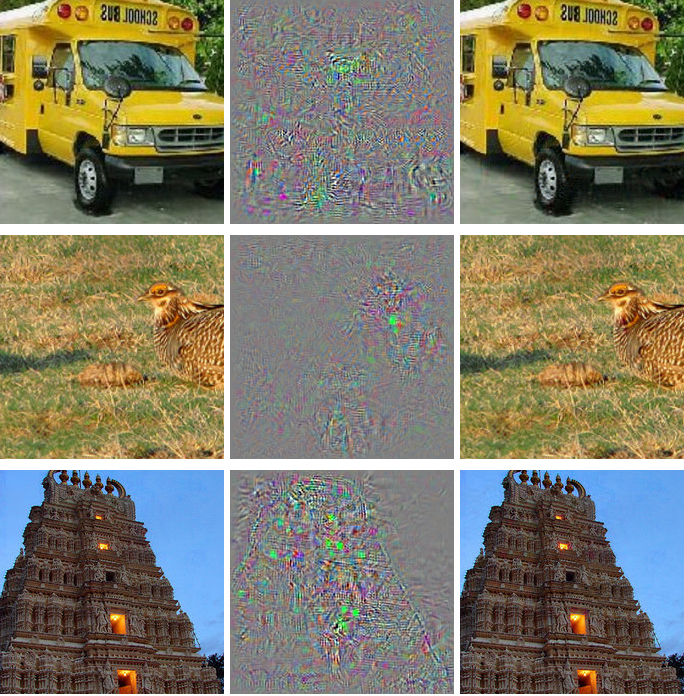}
  }
  \hspace{10mm}
  \subfigure[]{
        \includegraphics[width=0.32\textwidth]{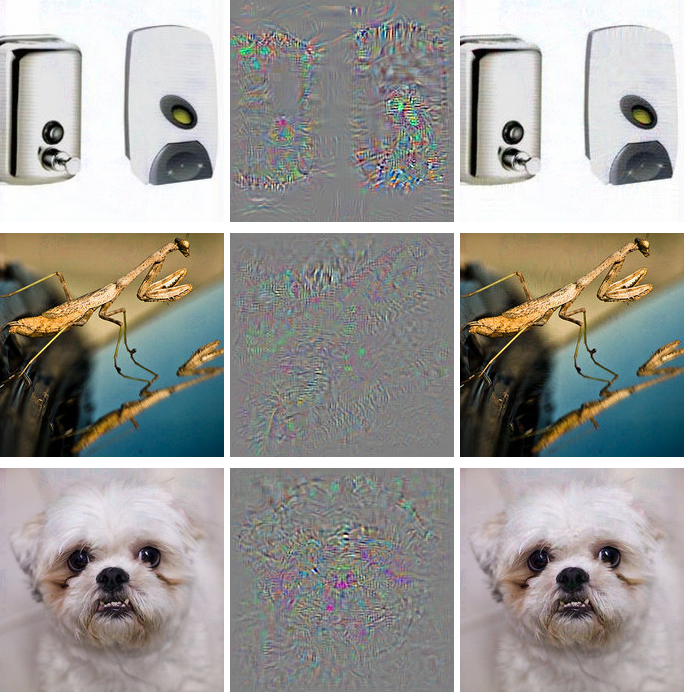}
  }
  \caption{Adversarial examples generated for AlexNet~\cite{krizhevsky2012imagenet}.(Left) is a correctly predicted sample, (center) difference between correct image, and image predicted incorrectly magnified by 10x (values shifted by 128 and clamped), (right) adversarial example. All images in the right column are predicted to be an {\it ``ostrich, Struthio camelus''}. Average distortion based on 64 examples is 0.006508. Plase refer to {\tt http://goo.gl/huaGPb} for full resolution images. The examples are strictly randomly chosen. There is not any postselection involved.}
  \label{fig:alexnegative}
\end{figure}
\begin{figure}
  \centering
  \subfigure[]{
        \includegraphics[width=0.35\textwidth]{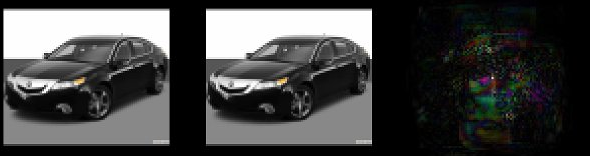}
  }
  \hspace{10mm}
  \subfigure[]{
        \includegraphics[width=0.35\textwidth]{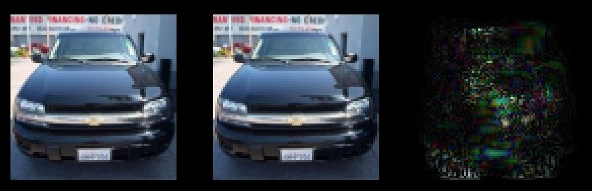}
  }
  \caption{Adversarial examples for QuocNet \cite{le2011building}.
A binary car classifier was trained on top of the last layer features without fine-tuning. 
The randomly chosen examples on the left are recognized correctly as cars,
while the images in the middle are not recognized.
The rightmost column is the magnified absolute value of the difference between
the two images.}
  \label{fig:quocnegative}
\end{figure}

Our ``minimimum distortion'' function $D$ has the following intriguing 
properties which we will support by informal evidence and quantitative
experiments in this section:
\begin{enumerate}
\item For all the networks we studied (MNIST, QuocNet \cite{le2011building}, 
AlexNet \cite{krizhevsky2012imagenet}), for each
sample, we have always managed to generate very close, visually hard to
distinguish, adversarial examples that are misclassified by the original
network (see figure \ref{fig:alexnegative} and {\tt http://goo.gl/huaGPb}
for examples).
\item {\it Cross model generalization:} a relatively large fraction of examples will
be misclassified by networks trained from scratch with different
hyper-parameters (number of layers, regularization or initial weights).
\item {\it Cross training-set generalization} a relatively large fraction of examples
will be misclassified by networks trained from scratch {\it on a disjoint
training set}.
\end{enumerate}
The above observations suggest that adversarial examples are somewhat universal 
and not just the results of overfitting to a particular model or to the specific
selection of the training set. They also suggest that back-feeding
adversarial examples to training might improve generalization of the resulting models.
Our preliminary experiments have yielded positive evidence on MNIST to
support this hypothesis as well: We have successfully trained a two layer
100-100-10 non-convolutional neural network with a test error below $1.2\%$ by
keeping a pool of adversarial examples a random subset of which is continuously
replaced by newly generated adversarial examples and which is mixed into the
original training set all the time. We used weight decay, but no dropout for
this network. For comparison, a network of this size gets to $1.6\%$ errors
when regularized by weight decay alone and can be improved to around $1.3\%$ 
by using carefully applied dropout. A subtle, but essential
detail is that we only got improvements by generating adversarial examples for
each layer outputs which were used to train all the layers above.
The network was trained in an alternating fashion, maintaining and updating
a pool of adversarial examples for each layer separately in addition to the
original training set. According to our initial observations, adversarial
examples for the higher layers seemed to be significantly more useful than
those on the input or lower layers. In our future work, we plan to compare these
effects in a systematic manner.

For space considerations, we just present results for a representative subset
(see Table \ref{crossneg}) of the MNIST experiments we performed.
The results presented here are consistent with those on a larger variety of
non-convolutional models. For MNIST, we do not have results for convolutional
models yet, but our first qualitative experiments with AlexNet gives us
reason to believe that convolutional networks may behave similarly as well.
Each of our models were trained with L-BFGS until convergence. The
first three models are linear classifiers that work on the pixel level
with various weight decay parameters $\lambda$.
All our examples use quadratic weight decay on the connection weights:
$\textrm{loss}_{decay}=\lambda\sum w_i^2/k$ added to the total loss, where $k$ is the 
number of units in the layer. Three of our models are simple linear (softmax)
classifier without hidden units (FC10($\lambda$)). One of them, 
FC10($1$), is trained with extremely high $\lambda=1$ in order to test whether 
it is still possible to generate adversarial examples in this extreme setting as 
well.Two other models are a simple sigmoidal neural network with two hidden layers 
and a classifier. The last model, AE400-10,
consists of a single layer sparse autoencoder with sigmoid
activations and 400 nodes with a Softmax classifier.
This network has been trained until it got very high quality first layer
filters and this layer was {\bf not} fine-tuned. The last column measures
the minimum average pixel level distortion necessary to reach $0\%$ accuracy
on the training set.
The distortion is measure by $\sqrt{\frac{\sum (x_i' - x_i)^2}{n}}$ between
the original $x$ and distorted $x'$ images, where $n=784$ is the number of
image pixels. The pixel intensities are scaled to be in the range $[0,1]$.

\begin{figure}
  \centering
  \subfigure[Even columns: adversarial examples for a linear (FC) classifier (stddev=0.06)]{
        \includegraphics[width=0.23\textwidth]{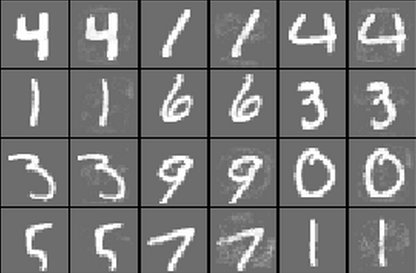}
  }
  \hspace{9mm}
  \subfigure[Even columns: adversarial examples for a 200-200-10 sigmoid network (stddev=0.063)]{
        \includegraphics[width=0.23\textwidth]{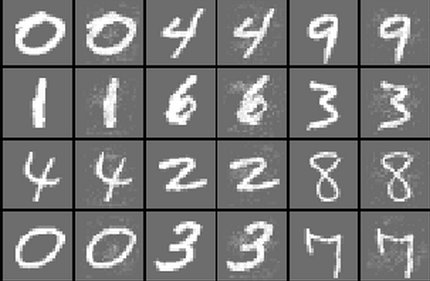}
  }
  \hspace{9mm}
  \subfigure[Randomly distorted samples by Gaussian noise with stddev=1.
             Accuracy: 51\%.]{
        \includegraphics[width=0.23\textwidth]{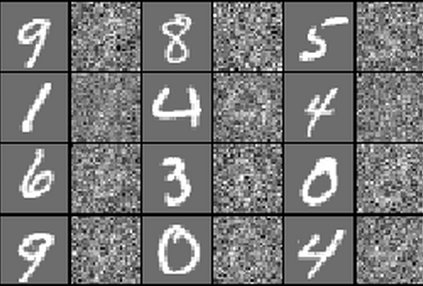}
  }
  \caption{Adversarial examples for a randomly chosen subset of MNIST compared with 
    randomly distorted examples. Odd columns correspond to original images, 
    and even columns correspond to distorted counterparts. The adversarial examples 
    generated for the specific model have accuracy 0\% for the respective model. 
    Note that while the randomly distorted examples are hardly readable, 
    still they are classified correctly in half of the cases, while the 
    adversarial examples are never classified correctly. }
  \label{fig:mnistdistorted}
\end{figure}

In our first experiment, we generated a set of adversarial instances for a given
network and fed these examples for each other network to measure the
proportion of misclassified instances. The last column shows the average
minimum distortion that was necessary to reach 0\% accuracy on the
whole training set. The experimental results are presented in
Table \ref{negativegen}. The columns of Table  \ref{negativegen} show the
error (proportion of misclassified instances) on the so distorted training
sets. The last two rows are given for reference showing the error induced
when distorting by the given amounts of Gaussian noise. Note that even
the noise with stddev 0.1 is greater than the stddev of our adversarial noise
for all but one of the models.
Figure \ref{fig:mnistdistorted} shows a visualization of the generated
adversarial instances for two of the networks used in this experiment
The general conclusion is that adversarial examples tend to stay hard
even for models trained with different hyperparameters.
Although the autoencoder based version seems most resilient to
adversarial examples, it is not fully immune either.

\begin{table}
\tiny
\centering
\begin{tabular}{|l||l|l|l|l|}
\hline
Model Name & Description & Training error & Test error & Av. min. distortion \\
\hline\hline
FC10($10^{-4}$) & Softmax with $\lambda=10^{-4}$ & 6.7\% & 7.4\% & 0.062 \\
\hline
FC10($10^{-2}$) & Softmax with $\lambda=10^{-2}$ & 10\% & 9.4\% & 0.1 \\
\hline
FC10($1$) & Softmax with $\lambda=1$ & 21.2\% & 20\% & 0.14 \\
\hline
FC100-100-10 & Sigmoid network $\lambda=10^{-5},10^{-5},10^{-6}$ & 0\% & 1.64\% & 0.058 \\
\hline
FC200-200-10 & Sigmoid network $\lambda=10^{-5},10^{-5},10^{-6}$ & 0\% & 1.54\% & 0.065 \\
\hline
AE400-10 & Autoencoder with Softmax $\lambda=10^{-6}$ & 0.57\% & 1.9\% & 0.086 \\
\hline
\end{tabular}
\caption{Tests of the generalization of adversarial instances on MNIST.}
\label{crossneg}
\end{table}

\begin{table}
\tiny
\centering
\begin{adjustwidth}{-1in}{-1in}
\begin{center}
\begin{tabular}{|l||l|l|l|l|l|l||c|}
\hline
& FC10($10^{-4}$) & FC10($10^{-2}$) & FC10($1$) & FC100-100-10 & FC200-200-10 & AE400-10 & Av. distortion  \\ 
\hline
FC10($10^{-4}$) & 100\% & 11.7\% & 22.7\% & 2\% & 3.9\% & 2.7\% & 0.062 \\
FC10($10^{-2}$) & 87.1\% & 100\% & 35.2\% & 35.9\% & 27.3\% & 9.8\% & 0.1 \\
FC10($1$)      & 71.9\% & 76.2\% & 100\% & 48.1\% & 47\% & 34.4\% & 0.14 \\ 
FC100-100-10 & 28.9\% & 13.7\% & 21.1\% & 100\% & 6.6\% & 2\% & 0.058 \\
FC200-200-10 & 38.2\% & 14\% & 23.8\% & 20.3\% & 100\% & 2.7\% & 0.065 \\
AE400-10     & 23.4\% & 16\% & 24.8\% & 9.4\% & 6.6\% & 100\% & 0.086 \\
\hline
Gaussian noise, stddev=0.1 & 5.0\%  & 10.1\% & 18.3\% & 0\% & 0\%   & 0.8\% & 0.1\\
Gaussian noise, stddev=0.3 & 15.6\% & 11.3\% & 22.7\% & 5\% & 4.3\% & 3.1\% & 0.3 \\
\hline
\end{tabular}
\end{center}
\end{adjustwidth}
\caption{Cross-model generalization of adversarial examples. The columns of the Tables show the error 
induced by distorted examples fed to the given model. The last column shows
average distortion wrt. original training set.}
\label{negativegen}
\end{table}

Still, this experiment leaves open the question of dependence over the training
set. Does the hardness of the generated examples rely solely on the particular
choice of our training set as a sample or does this effect generalize even to
models trained on completely different training sets?

\begin{table}
\tiny
\centering
\begin{tabular}{|l||l|l|l|l|}
\hline
Model & Error on $P_1$ & Error on $P_2$ & Error on Test & Min Av. Distortion \\
\hline\hline
FC100-100-10: 100-100-10 trained on $P_1$  & 0\%   & 2.4\% & 2\%   & 0.062 \\
\hline
FC123-456-10: 123-456-10 trained on $P_1$ & 0\%   & 2.5\% & 2.1\% & 0.059 \\
\hline
FC100-100-10' trained on $P_2$ & 2.3\% & 0\%   & 2.1\% & 0.058 \\
\hline
\end{tabular}
\caption{Models trained to study cross-training-set generalization of the 
         generated adversarial examples. Errors presented in Table correpond to original not-distorted data,
	 to provide a baseline.}
\label{crosstrainneg}
\end{table}

\begin{table}
\tiny
\centering
\begin{tabular}{|l||l|l|l|}
\hline
 & FC100-100-10 & FC123-456-10 & FC100-100-10' \\
\hline
Distorted for FC100-100-10  (av. stddev=0.062) & 100\%  & 26.2\% & 5.9\% \\
Distorted for FC123-456-10  (av. stddev=0.059) & 6.25\% & 100\%  & 5.1\% \\
Distorted for FC100-100-10' (av. stddev=0.058) & 8.2\% & 8.2\%   & 100\% \\
\hline
Gaussian noise with stddev=$0.06$ & 2.2\% & 2.6\% & 2.4\% \\
\hline\hline
Distorted for FC100-100-10 amplified to stddev=$0.1$ & 100\% & 98\%  & 43\% \\
Distorted for FC123-456-10 amplified to stddev=$0.1$ & 96\%  & 100\% & 22\% \\
Distorted for FC100-100-10' amplified to stddev=$0.1$ & 27\% & 50\%   & 100\% \\
\hline
Gaussian noise with stddev=$0.1$ & 2.6\% & 2.8\% & 2.7\% \\
\hline
\end{tabular}

\caption{Cross-training-set generalization error rate for the set of
adversarial examples generated for different models.
The error induced by a random distortion to the same
examples is displayed in the last row.}
\label{crosstrainnegresults}
\end{table}

To study cross-training-set generalization, we have partitioned
the 60000 MNIST training images into two parts $P_1$ and $P_2$ of size 30000
each and trained three non-convolutional networks with sigmoid activations on them: 
Two, FC100-100-10 and FC123-456-10, on $P_1$ and FC100-100-10 on $P_2$.
The reason we trained two networks for $P_1$ is to study the
cumulative effect of changing the hypermarameters and the training sets at
the same time. Models FC100-100-10 and FC100-100-10 share the same
hyperparameters: both of them are 100-100-10 networks, while FC123-456-10 has
different number of hidden units. In this experiment, we were distorting the 
elements of the test set rather than the training set. 
Table \ref{crosstrainneg} summarizes the
basic facts about these models. After we generate adversarial examples with $100\%$
error rates with minimum distortion for the test set, we feed these examples to
the each of the models. The error for each model is displayed in the corresponding 
column of the upper part of Table \ref{crosstrainnegresults}.
In the last experiment, we magnify the effect of our distortion by using the
examples  $x + 0.1\frac{x' - x}{\|x' - x\|_2}$ rather than $x'$. This magnifies the
distortion on average by 40\%, from stddev $0.06$ to $0.1$. The so distorted examples
are fed back to each of the models and the error rates are displayed in the
lower part of Table \ref{crosstrainnegresults}. The intriguing conclusion is
that the adversarial examples remain hard for models trained even on a disjoint
training set, although their effectiveness decreases considerably.

\subsection{Spectral Analysis of Unstability}\label{sec:param}

The previous section showed examples 
of deep networks resulting from 
purely supervised training 
which are unstable with respect to a peculiar 
form of small perturbations. 
Independently of their generalisation properties across 
networks and training sets, the adversarial examples 
show that there exist small additive perturbations of the input (in Euclidean sense) 
that produce large perturbations at the output of the last layer.
This section describes a simple procedure to 
measure and control the additive stability of the network 
by measuring the spectrum of each rectified layer.

Mathematically, if $\phi(x)$ denotes the output of a network 
of $K$ layers corresponding to input $x$ and trained parameters $W$, 
we write 
$$\phi(x) = \phi_K( \phi_{K-1}( \dots \phi_1(x; W_1) ; W_2 )\dots ; W_K)~,$$
where $\phi_k$ denotes the operator mapping layer $k-1$ to layer $k$. 
The unstability of $\phi(x)$ can be explained by inspecting the upper Lipschitz constant 
of each layer $k=1\dots K$, defined as the constant $L_k>0$ such that
$$\forall\, x,\,r~,~ \| \phi_k(x; W_k) - \phi_k(x+r; W_k) \| \leq L_k \| r \|~.$$
The resulting network thus satsifies $\| \phi(x) - \phi(x+r) \| \leq L \| r \|$, with
$L = \prod_{k=1}^K L_k$.

A half-rectified layer (both convolutional or fully connected) is defined by the mapping 
$\phi_k(x; W_k,b_k) = \max(0, W_k x+b_k)$. 
Let $\|W \|$ denote the operator norm of $W$ (i.e., its largest singular value). 
Since the non-linearity $\rho(x) = \max(0,x)$ is contractive, i.e. satisfies 
$\| \rho(x) - \rho(x+r) \| \leq \| r\|~$ for all $x,r$;  it follows that  
 $$ \| \phi_k(x; W_k) - \phi_k(x+r; W_k) \| = \| \max(0,W_k x + b_k) - \max(0,W_k (x+r) + b_k) \| \leq \| W_k r \| \leq \|W_k \| \| r\|~,$$
 and hence $L_k \leq \|W_k\|$. 
On the other hand, a max-pooling layer $\phi_k$ is contractive: 
$$\forall\, x\,,\,r\,,~\| \phi_k(x) - \phi_k(x+r) \| \leq \| r\| ~,$$
since its Jacobian is a projection onto a subset of the input coordinates 
and hence does not expand the gradients.
Finally, if $\phi_k$ is a contrast-normalization layer
$$\phi_k(x) = \frac{x}{\Big( \epsilon + \|x \|^2 \Big)^\gamma}~,$$
one can verify that 
$$\forall\, x\,,\,r\,,~\| \phi_k(x) - \phi_k(x+r) \| \leq  \epsilon^{-\gamma} \| r\| $$
for $\gamma \in [0.5,1]$, which corresponds to most common operating 
regimes.

It results that a conservative measure of the 
unstability of the network can be obtained by simply computing 
the operator norm of each fully connected and convolutional layer. 
The fully connected case is trivial since the norm is directly given 
by the largest singular value of the fully connected matrix.
Let us describe the convolutional case.
If $W$ denotes a generic $4$-tensor, implementing a convolutional layer with 
$C$ input features, $D$ output features, support $N\times N$ and spatial stride $\Delta$, 
$$W x = \left\{  \sum_{c=1}^C x_c \star w_{c,d} (n_1 \Delta, n_2 \Delta) \, ; d=1\,\dots,D\right\}~,$$
where $x_c$ denotes the $c$-th input feature image, 
and $w_{c,d}$ is the spatial kernel corresponding to input feature $c$ and output feature $d$,
by applying Parseval's formula we obtain that
its operator norm is given by
\begin{equation}
\label{convbound}
\| W \| = \sup_{\xi \in [0,N \Delta^{-1})^2} \| A(\xi) \| ~,
\end{equation}
where $A(\xi)$ is a $D \times (C \cdot \Delta^{2})$ matrix whose rows are 
$$\forall~d=1\dots D~,~A(\xi)_d =  \Big(\Delta^{-2}  \widehat{w_{c,d}}(\xi+ l \cdot N \cdot \Delta^{-1}) \,;\, c=1\dots C\, ,\, l=(0\dots \Delta-1)^2 \Big)~,$$
and $\widehat{w_{c,d}}$ is the 2-D Fourier transform of $w_{c,d}$:
$$\widehat{w_{c,d}}(\xi) = \sum_{u \in [0,N)^2} w_{c,d}(u) e^{- 2 \pi i (u \cdot \xi) / N^2}~.$$

\begin{table}
\tiny
\centering
\begin{tabular}{|c| c  c | c |}
\hline
Layer & Size & Stride & Upper bound\\
\hline
Conv. $1$ & $3 \times 11 \times 11 \times 96$ & $4$  & $2.75$ \\
Conv. $2$ & $96 \times 5 \times 5 \times 256$ & $1$ & $10$ \\
Conv. $3$ & $256 \times 3 \times 3 \times 384$ & $1$ & $7$ \\
Conv. $4$ & $384 \times 3 \times 3 \times 384$ & $1$ & $7.5$ \\
Conv. $5$ & $384 \times 3 \times 3 \times 256$ & $1$ & $11$ \\
\hline
FC. 1 & $9216 \times 4096$ & N/A&  $3.12$ \\
FC. 2 & $4096 \times 4096$ & N/A &  $4$ \\
FC. 3 & $4096 \times 1000$ & N/A &  $4$ \\
\hline
\end{tabular}
\caption{Frame Bounds of each rectified layer of the network from  \cite{krizhevsky2012imagenet}.}
\label{bounds}
\end{table}

Table \ref{bounds} shows the upper Lipschitz bounds computed from the ImageNet deep 
convolutional network of \cite{krizhevsky2012imagenet}, using (\ref{convbound}).
 It shows that instabilities can appear as soon as in the first
convolutional layer. 

These results are consistent with the exsitence of blind spots 
constructed in the previous section, but they don't attempt to explain why
these examples generalize across different hyperparameters or training sets.
We emphasize that we compute upper bounds: 
large bounds do not automatically translate into existence of adversarial examples;
however, small bounds guarantee that no such examples can appear.  
This suggests a simple regularization of the parameters, 
consisting in penalizing each upper Lipschitz bound, 
which might help improve the generalisation error of the networks.

\section{Discussion}

We demonstrated that deep neural networks have counter-intuitive properties both with respect to the semantic meaning of individual units and with respect to their discontinuities.  The existence of the adversarial negatives appears to be in contradiction with the network's ability
to achieve high generalization performance. Indeed, if the network can generalize well, how can it be confused by these adversarial negatives, which are indistinguishable from the regular examples?  Possible explanation is that the set of adversarial negatives is of extremely low probability, and thus is never (or rarely) observed in the test set, yet it is dense (much like the rational numbers), and so it is found near every virtually every test case. However, we don't have a deep understanding of how often adversarial negatives appears, and thus this issue should be addressed in a future research.


\bibliographystyle{plain}

\end{document}